%% file: main.tex
\documentclass[runningheads]{llncs}
\usepackage[T1]{fontenc}
\IfFileExists{newtxtext.sty}{%
  \usepackage{newtxtext}\usepackage[varvw]{newtxmath}}{%
  \usepackage{mathptmx}}
\usepackage{amsmath}
\usepackage{bm}
\usepackage{graphicx}
\usepackage{booktabs}
\usepackage{multirow}
\usepackage{xcolor}
\usepackage[capitalize,nameinlink]{cleveref}
\usepackage{url}
\newcommand{\R}{\mathbb{R}}
\newcommand{\Sph}{\mathbb{S}}
\newcommand{\SO}{\mathit{SO}(3)}
\newcommand{\Keff}{K_{\mathrm{eff}}}
\newcommand{\Hn}{\hat{H}}
\crefname{equation}{Eq.}{Eqs.}
\crefname{figure}{Fig.}{Figs.}
\crefname{table}{Table}{Tables}
\crefname{section}{Sect.}{Sects.}

\begin{document}
\title{Does Pose Encoding Matter? Analyzing Geometric Representations for Discrete Motion Vocabulary Learning}
\titlerunning{Does Pose Encoding Matter?}
\author{Amir Kimiyaie\inst{1} \and
Nahal Mafi\inst{1} \and
Hamed Tabkhi\inst{1}}
\authorrunning{A. Kimiyaie et al.}
\institute{Department of Electrical and Computer Engineering,\\
University of North Carolina Charlotte, Charlotte NC 28223, USA\\
\email{\{akimiyai,nmafi,htabkhiv\}@charlotte.edu}}
\maketitle
\begin{abstract}
Compressing continuous human motion into a finite vocabulary of discrete tokens is now the first stage of many motion understanding and generation pipelines. Vector quantization partitions a latent space into Voronoi cells around a finite codebook, but the geometry of that latent space is inherited from the input pose encoding, and it is unclear which encoding admits an efficient discretization. We study four encodings that place skeletal motion in different geometric spaces: Gaussian joint heatmaps on the pixel lattice, which we show to be a morphological dilation of the joint point set by a quadratic structuring function; root-relative joint positions in Euclidean space; local joint rotations on $\SO$ through a continuous 6D embedding; and a redundant full kinematic encoding. For each, we train VQ-VAE tokenizers whose codebooks lie on the unit hypersphere, varying codebook size, temporal stride, and code dimension. Beyond reconstruction, we measure codebook utilization and the effective vocabulary size, i.e., the number of equiprobable spherical Voronoi cells with the same entropy as the observed token distribution. Rotation and position encodings keep converting additional codewords into effective vocabulary, whereas the full kinematic encoding saturates at a few hundred effective cells and heatmaps at fewer than sixty. The smallest code dimension gives the highest utilization for three of the four encodings, and the effect of temporal stride depends on the encoding. Encodings of the same pose family can thus yield effective vocabularies differing by an order of magnitude, a difference that reconstruction loss alone does not reveal, which points to input geometry as a first-order design choice in motion tokenization.

\keywords{Discrete motion vocabulary \and Vector quantization \and Spherical Voronoi tessellation \and Rotation group \and Morphological dilation \and Geometric deep learning}
\end{abstract}
\input{sections/intro}
\input{sections/related}
\input{sections/input_repr}
\input{sections/vectorquant}
\input{sections/experiments}

\section{Conclusion}
\label{sec:conclusion}
Viewing the motion tokenizer as a spherical Voronoi tessellation and counting how many of its cells are effectively used, we found that the geometric encoding of pose matters more than reconstruction loss suggests. We draw three observations that, to our knowledge, have not been reported for motion tokenization. First, \emph{the same intrinsic dimension does not imply the same usable vocabulary}. Three encodings of the same $(\Sph^2)^{21}$ pose family differ by more than an order of magnitude in effective vocabulary: the continuous rotation embedding makes the best use of the tessellation, while the redundant full kinematic encoding saturates at a few hundred cells. Second, \emph{reconstruction loss and vocabulary usage can disagree}, so motion tokenizers should be reported with utilization and $\Keff$ alongside reconstruction. Third, \emph{lattice heatmaps use few cells}: this encoding, a morphological dilation of the digitized joint set that is discretized twice, used fewer than sixty effective cells when varying $K$. Low code dimensions were generally preferable, consistent with latents that occupy a low-dimensional part of the sphere, while the best temporal stride depended on the encoding.

These observations give practical guidance for building motion tokenizers, and suggest that discrete geometry and morphology can help explain, not only describe, learned discrete representations. Our evidence has clear limits. Each configuration was trained once, on one dataset, and the encodings also differ in training details. The heatmap model uses a transformer and fewer epochs, and the full kinematic model uses a larger batch and an auxiliary position loss. We therefore read the gaps between encodings as strong indications of a geometric effect rather than as isolated causal estimates. Controlled retraining with matched architectures, batch sizes, and losses, repeated over several seeds, is the most direct next step. We will also relate effective vocabulary to downstream performance and analyze the learned codebooks as spherical point sets (packing and covering radii, relation to action labels).


\bibliographystyle{splncs04}
\bibliography{references}
\end{document}

%% file: sections/intro.tex
\section{Introduction}
\label{sec:intro}

The human body is naturally modeled as an articulated skeleton: a kinematic tree of rigid segments whose motion is determined by the rigid transformation of each joint relative to its parent. Action recognition, anomaly detection, gait analysis, and motion synthesis therefore operate on skeletal pose sequences, obtained from motion capture~\cite{AMASS:ICCV:2019} or from 2D/3D pose estimation~\cite{alphapose,8237550}. The same underlying geometry is, however, parameterized in very different spaces: as points digitized on a pixel lattice, as point sets in $\R^3$, or as elements of the rotation group $\SO$. Each space has its own continuity, redundancy, and constraint structure.

A widely adopted way to learn from such data is \emph{motion tokenization}~\cite{momask,t2m-gpt}, which is at heart a discretization problem: continuous trajectories are mapped to words of a finite alphabet. A vector-quantized autoencoder (VQ-VAE)~\cite{vq-vae} sends each short motion segment to the nearest element of a learned codebook, and the resulting token sequences are modeled by autoregressive or masked transformers. Geometrically, the quantizer partitions the latent space into the Voronoi cells of the codebook~\cite{gersho1992vector}. When latents and codewords are $\ell_2$-normalized, as in our models, this partition is a tessellation of the unit hypersphere. Downstream models only see which cell was hit, so the quality of this discretization bounds everything built on top of it.

\textbf{Motivation.} The tokenizer is not a detail. It is trained first, then frozen, and it fixes the vocabulary on which generative, captioning, and recognition models are built~\cite{momask,t2m-gpt,TM2T}; any motion it cannot name is lost to every model trained on its tokens. Yet its most basic design choice, the input encoding, is usually made by convention or by the data source: heatmaps when poses come from an online 2D estimator, rotations when they come from motion capture, the full kinematic feature set because popular generation models use it. Codebook size, code dimension, and temporal stride are then inherited from image and audio models, and the tokenizer is judged by reconstruction loss alone. This first stage has received little systematic study for human motion~\cite{oliveira2026benchmarking}, and it is unclear whether encodings that describe the same skeleton lead to vocabularies of comparable quality. Geometry suggests they may not. The encoder must map every input space onto the same sphere, so discontinuities of the parameterization~\cite{zhou2019continuity}, unenforced constraints such as fixed bone lengths, and redundant channels can all change how the Voronoi cells of the codebook are used.

We study this question from the viewpoint of discrete geometry and mathematical morphology. Our contributions are:
\begin{itemize}
  \item A geometric account of four skeletal encodings (\cref{sec:encodings}). We show that max-merged Gaussian heatmaps are the dilation of the joint set by a quadratic structuring function (\cref{prop:heatmap}), and that rotations recovered by minimal-twist inverse kinematics carry exactly the 42 degrees of freedom of bone directions.
  \item A formulation of motion quantization as a spherical Voronoi tessellation, together with capacity measures, namely utilization and effective vocabulary size, that count how many cells a tokenizer actually uses (\cref{sec:vq-vae}).
  \item A controlled study over codebook size, temporal stride, and code dimension, showing that the choice of encoding strongly affects whether additional codewords become usable vocabulary (\cref{sec:experiments}).
\end{itemize}

\textbf{Main insight.} Encodings with the same intrinsic degrees of freedom need not yield vocabularies of the same size. Positions, rotations, and the full kinematic encoding all describe the same 42-dimensional family of poses (\cref{tab:spaces}). Yet at $K=8192$ their tokenizers use about $3000$, $5200$, and $380$ effective codewords, respectively, while the full kinematic tokenizer's reconstruction loss barely changes with $K$. This points to the \emph{parameterization} of the pose, i.e., its continuity, redundancy, and digitization, as a major factor in how much of a discrete vocabulary becomes usable, a factor that reconstruction loss alone does not reveal. Because the encodings also differ in some training details, we treat this as a strong empirical indication and discuss the controls needed to isolate it (\cref{sec:conclusion}). This gives practitioners a concrete criterion for choosing encodings and codebook sizes, and shows how Voronoi tessellations, morphology, and manifold structure can help explain modern learned models.

%% file: sections/related.tex
\section{Related Work}
\label{sec:related}

\textbf{Quantization geometry and morphology.}
A nearest-neighbor quantizer partitions its input space into the Voronoi cells of its codebook, and Lloyd's algorithm alternates between this partition and moving each codeword to the centroid of its cell~\cite{gersho1992vector}. With unit-norm data, cosine similarity replaces distance and the centroid step becomes a normalized mean, which gives spherical $k$-means~\cite{dhillon2001concept}. In mathematical morphology, dilation by a quadratic structuring function is the morphological counterpart of Gaussian smoothing and generates a morphological scale-space~\cite{vandenboomgaard1994morphological,serra1982image}. We use both views to analyze motion tokenizers.

\textbf{Pose representations.}
Skeletal motion is studied in geometric deep learning~\cite{geometric-rigid-nonrigid}, vision~\cite{alphapose}, robotics~\cite{7506114}, and graphics~\cite{motionbricks}, using either Cartesian joint coordinates or joint rotations combined with kinematics. For rotations the parameterization matters. Zhou et al.~\cite{zhou2019continuity} proved that every representation of $\SO$ in a Euclidean space of dimension at most four is discontinuous and proposed continuous 6D and 5D alternatives, and Geist et al.~\cite{3692070.3692684} recommend $\R^9$ with SVD or $\R^6$ with Gram--Schmidt for general use. Joint heatmaps~\cite{9879048,baharani2026feasibility} rasterize joints on a pixel grid to absorb coordinate noise, and frequency encodings~\cite{gsps,10.1007/978-3-030-58571-6_14} represent smooth trajectories by truncated discrete cosine transforms.

\textbf{Discrete motion representations.}
Learning from pose sequences underpins much of human behavior analysis~\cite{9795869,11322811}. Discrete VAEs~\cite{dalle} and VQ-VAEs~\cite{vq-vae} learn a finite codebook that serves as a vocabulary for sequence models conditioned on text~\cite{TM2T}, audio~\cite{bailando}, or images~\cite{tokenhmr}. In motion generation, VQ-VAE tokenizers combined with masked~\cite{momask} or discrete-diffusion~\cite{10377752} transformers are standard, and residual~\cite{momask} and multi-scale~\cite{snapmogen} quantization improve reconstruction. Codebooks tend to collapse onto few codewords, which has motivated moving-average updates, codeword resets~\cite{t2m-gpt}, and finite scalar quantization~\cite{ICLR2024_e2dd5360}. A recent benchmark compares motion tokenizers under controlled settings~\cite{oliveira2026benchmarking}. However, the joint effect of the input encoding's geometry and of codebook size, code dimension, and temporal stride on how the quantizer's cells are used has not been studied.

%% file: sections/input_repr.tex
\section{Geometric Encodings of Skeletal Motion}
\label{sec:encodings}

\begin{figure}[t]
  \centering
  \includegraphics[width=\linewidth]{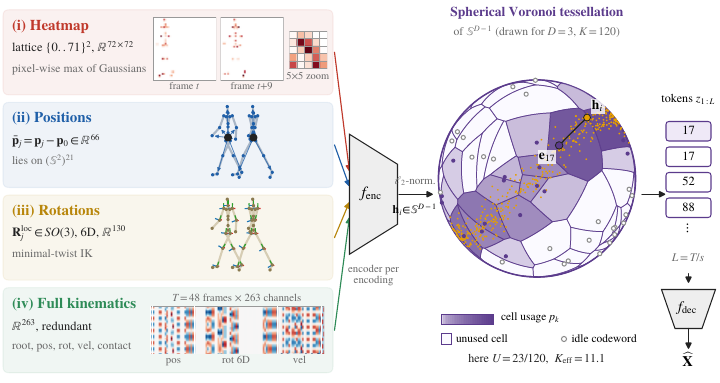}
  \Description{Left: four rows, one per encoding, computed from the same walking skeleton. (i) Two 72 by 72 heatmap frames whose joints appear as small dots, with a 5 by 5 zoom showing that each joint covers about a 3 by 3 pixel neighborhood. (ii) Two skeletons with arrows from the pelvis to the joints. (iii) The same skeletons with small red, green and blue local coordinate frames at the joints. (iv) A 48 by 263 colored feature matrix split into position, rotation and velocity blocks. Arrows from all four rows enter a shared encoder. Its normalized output lies on a sphere partitioned into 120 spherical Voronoi cells. Orange latents cluster along a curve, cells are shaded by how often they are used, and most cells are empty with hollow, idle codewords. One latent is linked to its nearest codeword, number 17, which becomes the first token of the sequence passed to the decoder.}
  \caption{Overview, drawn from one synthetic walking sequence using the definitions of \cref{sec:encodings}. Left: the four encodings, (i)~max-merged Gaussian heatmaps on a $72\times72$ lattice (the zoom shows the ${\approx}3\times3$ support of one joint), (ii)~pelvis-relative positions, (iii)~local rotations from minimal-twist IK, drawn as joint frames, and (iv)~the $T\times263$ full kinematic feature matrix. Right: normalized latents $\mathbf{h}_i$ (orange) are assigned to the nearest codeword, i.e., to a cell of the spherical Voronoi tessellation, and the cell index is the token. Shading shows how often each cell is used. Because the latents concentrate near a curve, most cells stay empty: here 23 of 120 cells are used and $\Keff=11.1$ (\cref{sec:vq-vae}).}
  \label{fig:schematic}
\end{figure}

We compare four encodings of skeletal motion that are common in practice and that describe the same motion in different spaces (\cref{fig:schematic,tab:spaces}): (i)~2D joint heatmaps on a pixel lattice, (ii)~root-relative 3D joint positions, (iii)~local joint rotations in $\SO$, and (iv)~a full kinematic encoding that combines rotations, positions, velocities, and foot contacts.

\textbf{Data.}
Frames are indexed by $t$ and joints by $j$, with $j=0$ the pelvis and $\pi(j)$ the parent of $j$. Encodings (ii)--(iv) are derived from Motion-X~\cite{lin2023motionx} with $J=22$ joints, given directly or reconstructed from SMPL-X/SMPL-H parameters. Sequences are resampled to 20 frames per second (linear interpolation for positions, SLERP for rotations), expressed in a Y-up frame, and cut into windows of $T=48$ frames. Normalization statistics come from the training set only.

\subsubsection{(i) Heatmaps as a morphological dilation.}
Heatmaps are popular in online settings: drawing each noisy predicted joint as a small blob makes the input tolerant to small displacements. Following~\cite{baharani2026feasibility}, we take 2D joint coordinates $\mathbf{x}_{t,j}$, obtained with random rotations about the $z$-axis, and fit each skeleton into an $S\times S$ pixel lattice with $S=72$. Each joint becomes a Gaussian evaluated at the lattice points $\mathbf{u}$, and all joints share one channel through a pixel-wise maximum:
\begin{equation}
  \mathcal{H}_t(\mathbf{u})=\max_{j}\,\exp\!\Big(-\frac{\|\mathbf{u}-\mathbf{x}_{t,j}\|_2^2}{2\sigma^2}\Big),
  \qquad \sigma=0.6 .
  \label{eq:heatmap}
\end{equation}
Stacking frames gives the input volume $\mathcal{H}\in\R^{T\times S\times S}$.

The maximum is what links this encoding to mathematical morphology. The dilation of a function $f$ by a structuring function $b$ is $(f\oplus b)(\mathbf{u})=\sup_{\mathbf{y}}\{f(\mathbf{y})+b(\mathbf{u}-\mathbf{y})\}$~\cite{serra1982image}. Taking the logarithm of \cref{eq:heatmap} turns each Gaussian into an inverted paraboloid and leaves the maximum in place.

\begin{proposition}\label{prop:heatmap}
Let $X_t=\{\mathbf{x}_{t,j}\}_j$ be the joint set, $\iota_{X_t}$ the function that is $0$ on $X_t$ and $-\infty$ elsewhere, and $q_\sigma(\mathbf{u})=-\|\mathbf{u}\|_2^2/2\sigma^2$. Then
\begin{equation}
  \log\mathcal{H}_t=\iota_{X_t}\oplus q_\sigma ,
  \label{eq:dilation}
\end{equation}
and $\mathcal{H}_t\le1$, with equality exactly at the joints.
\end{proposition}
\begin{proof}
Since $\iota_{X_t}=-\infty$ off $X_t$, the supremum runs over the joints only and equals $\max_jq_\sigma(\mathbf{u}-\mathbf{x}_{t,j})$, which is the logarithm of \cref{eq:heatmap} because $\log$ is increasing. Each Gaussian is at most~1, with equality only at its own joint.\qed
\end{proof}

In words, the log-heatmap is the joint set dilated by a quadratic (parabolic) structuring function, and $\sigma$ acts as a morphological scale~\cite{vandenboomgaard1994morphological}. The second statement explains why the maximum is preferred over the sum: every joint keeps its peak, whereas summing two Gaussians closer than $2\sigma$ fuses them into one peak at their midpoint. With $\sigma=0.6$, a joint's value drops to $0.25$ one pixel away and $0.004$ two pixels away, so each joint occupies about a $3\times3$ neighborhood (\cref{fig:schematic}, zoom). The heatmap is therefore close to a \emph{digitization} of the joint set, and this encoding discretizes the pose twice: on the lattice, and again in latent space by the quantizer of \cref{sec:vq-vae}.

\subsubsection{(ii) Root-relative joint positions.}
Subtracting the pelvis from every joint, $\bar{\mathbf{p}}_{t,j}=\mathbf{p}_{t,j}-\mathbf{p}_{t,0}$, removes where the person is but keeps the body's configuration, orientation, and proportions, giving $\mathbf{X}^{\mathrm{pos}}\in\R^{T\times66}$. Features are centered per coordinate but divided by a single scale shared by all coordinates, namely the mean standard deviation. A common scale preserves the shape of the skeleton, whereas standardizing each coordinate separately would stretch the body differently along each axis. The 66 numbers are far from free. For a given subject the bone lengths are fixed, so a frame is determined by the 21 unit directions of its bones, i.e., by a point of $(\Sph^2)^{21}$, a 42-dimensional set inside $\R^{66}$ (\cref{tab:spaces}). The network is not told about this constraint and must learn it from data.

\begin{table}[!b]
  \centering
  \caption{Geometry of the four encodings: per-frame ambient space, the space in which the pose actually varies, and its per-frame degrees of freedom (dof). The root features of (iii) and (iv) are counted separately.}
  \label{tab:spaces}
  \small
  \setlength{\tabcolsep}{5pt}
  \begin{tabular}{@{}llll@{}}
    \toprule
    Encoding & Ambient & Underlying geometric space & dof \\
    \midrule
    (i) Heatmap      & $\R^{72\times72}$ & lattice samples of $\exp(\iota_X\oplus q_\sigma)$, $X\subset\R^2$, $|X|=22$ & $\le 44$ \\
    (ii) Positions   & $\R^{66}$         & $(\Sph^2)^{21}$ per subject (fixed bone lengths) & 42 \\
    (iii) Rotations  & $\R^{130}$        & $V_2(\R^3)^{21}\cong\SO^{21}$; IK image $\cong(\Sph^2)^{21}$ & $42+4$ \\
    (iv) Full kin.   & $\R^{263}$        & as (iii), plus redundant positions, velocities, contacts & $42+4$ \\
    \bottomrule
  \end{tabular}
\end{table}

\subsubsection{(iii) Local rotations on $\SO$.}
Rotations describe how each bone is turned relative to its parent rather than where the joints are. Since our source data mixes formats, we recover them from joint positions by inverse kinematics (IK) on a common skeleton whose bones have fixed rest offsets $\mathbf{o}_j$. They relate to positions through forward kinematics~\cite{PoseGPT,tokenhmr}:
\begin{equation}
  \mathbf{R}_{t,j}=\mathbf{R}_{t,\pi(j)}\mathbf{R}^{\mathrm{loc}}_{t,j},
  \qquad
  \mathbf{p}_{t,j}=\mathbf{p}_{t,\pi(j)}+\mathbf{R}_{t,j}\mathbf{o}_j .
  \label{eq:fk}
\end{equation}
The global rotation of a joint is thus the product of all local rotations from the pelvis to that joint, which is why a small change at the shoulder moves the whole arm. IK runs this recursion backwards. For each bone, we take the smallest rotation that turns its rest direction $\mathbf{o}_j/\|\mathbf{o}_j\|_2$ into its observed direction, and express it relative to the rotation already accumulated by the parent. A bone direction says nothing about the twist of the bone around itself, and this minimal-rotation choice has a simple consequence. The accumulated rotation of every joint equals its own alignment rotation, because the parent's rotation is first removed and then re-applied. Each local rotation therefore depends only on the directions of a bone and its parent. After IK, rotations carry exactly the 42 degrees of freedom of the positions in (ii), re-expressed on $\SO^{21}$ (\cref{tab:spaces}). The only singular configurations are bones pointing exactly opposite to their rest direction, where the minimal rotation is not unique.

Each local rotation matrix is fed to the network as its first two columns, a 6D vector from which the third column follows by a cross product. Unlike Euler angles, which suffer from gimbal lock, or quaternions, where $\mathbf{q}$ and $-\mathbf{q}$ denote the same rotation, this representation has no discontinuities~\cite{zhou2019continuity}. Together with four root features (heading angular velocity, horizontal root velocity, and root height), this gives $\mathbf{X}^{\mathrm{rot}}\in\R^{T\times130}$.

\subsubsection{(iv) Full kinematics.}
Our fourth encoding is the feature set widely used in motion generation~\cite{humanml3d,momask}. It augments (iii) with quantities that the rotations already determine but that a network may find easier to use directly. These are local joint positions (63 values, in the root-facing frame), joint velocities between consecutive frames (66 values), and four binary foot contacts, set when an ankle or toe moves less than a threshold ($\tau=0.002$) between frames. This gives $\mathbf{X}^{\mathrm{full}}\in\R^{T\times263}$, normalized with a standard deviation shared within each feature group so that no group dominates the loss. Positions and rotations are functions of the same bone directions and velocities are their differences, so the 263 channels re-encode the per-frame degrees of freedom of (iii) about $5.7$ times over. This redundancy is exactly what we want to test against quantization.

%% file: sections/vectorquant.tex
\section{Quantization as a Spherical Voronoi Tessellation}
\label{sec:vq-vae}

\textbf{Quantizer.}
An encoder $f_{\mathrm{enc}}$ maps a window $\mathbf{X}\in\R^{T\times F}$, in any of the four encodings, to a sequence of $L=T/s$ latent vectors, where the temporal stride $s$ is the number of frames summarized by one token. The latents are $\ell_2$-normalized, so each lies on the unit sphere, $\mathbf{h}_i\in\Sph^{D-1}$. The motion vocabulary is a set $\mathcal{V}=\{\mathbf{e}_k\}_{k=0}^{K-1}\subset\Sph^{D-1}$ of $K$ unit codewords, and every latent is replaced by its nearest codeword. On the sphere, $\|\mathbf{h}-\mathbf{e}\|_2^2=2-2\langle\mathbf{h},\mathbf{e}\rangle$, so nearest in distance means largest cosine similarity:
\begin{equation}
  z_i=\arg\max_k\,\langle\mathbf{h}_i,\mathbf{e}_k\rangle,
  \qquad
  V_k=\{\mathbf{h}\in\Sph^{D-1}:\langle\mathbf{h},\mathbf{e}_k-\mathbf{e}_l\rangle\ge0\ \forall l\}.
  \label{eq:vq_assignment}
\end{equation}
Each codeword $\mathbf{e}_k$ claims the cell $V_k$ of points closer to it than to any other codeword, and the cells $\{V_k\}$ form the spherical Voronoi tessellation of the codebook (\cref{fig:schematic}, right). A token $z_i$ is simply the name of the cell that contains $\mathbf{h}_i$, and the downstream sequence model sees only these names. The decoder reconstructs the window from the selected codewords, $\widehat{\mathbf{X}}=f_{\mathrm{dec}}(\mathbf{e}_{z_1},\dots,\mathbf{e}_{z_L})$. Because $\arg\max$ has no useful gradient, a straight-through estimator copies the decoder's gradient from $\mathbf{e}_{z_i}$ to $\mathbf{h}_i$.

The codebook is initialized by cosine $k$-means on training latents and then follows the data. Each codeword moves towards the mean of the latents currently in its cell, through an exponential moving average with decay $0.99$, and is renormalized onto the sphere. Alternating cell assignment~\eqref{eq:vq_assignment} with this update is an online version of Lloyd's algorithm on the sphere, i.e., spherical $k$-means~\cite{dhillon2001concept}. Its fixed points are centroidal spherical Voronoi tessellations of the latent distribution~\cite{gersho1992vector}.

\textbf{Objective.}
The training loss is
\begin{equation}
  \mathcal{L}=\mathcal{L}_{\mathrm{rec}}(\widehat{\mathbf{X}},\mathbf{X})
  +\lambda_{\mathrm{pos}}\,\mathcal{L}_{\mathrm{pos}}(\widehat{\mathbf{P}},\mathbf{P})
  +\frac{\beta}{LD}\sum_{i=1}^{L}\big\|\mathbf{h}_i-\mathrm{sg}[\mathbf{e}_{z_i}]\big\|_2^2 .
  \label{eq:vq_loss}
\end{equation}
The first term asks the decoder to reproduce the input (mean Smooth-$\ell_1$). The second applies the same loss to the 63 joint-position channels $\mathbf{P}$ of the full kinematic encoding, so that errors in its redundant channels cannot hide errors in the pose itself. It is used only for that encoding ($\lambda_{\mathrm{pos}}=0.5$, otherwise $0$). The last term, the commitment loss with $\beta=0.02$, pulls each latent towards the codeword of its cell. The stop-gradient $\mathrm{sg}[\cdot]$ makes this term move only the encoder, since codewords follow the moving average.

\textbf{How many cells are used?}
Reconstruction loss does not show whether the vocabulary is used well: a tokenizer can reconstruct reasonably while sending almost all latents to a few cells. Let $p_k$ be the empirical frequency with which cell $V_k$ is selected. We report three quantities: the \emph{utilization} $U=|\{k:p_k>0\}|/K$, the fraction of cells used at least once; the \emph{normalized entropy} $\Hn=H/\log K$, with $H=-\sum_kp_k\log p_k$; and the \emph{effective vocabulary size}
\begin{equation}
  \Keff=e^{H}=K^{\Hn},
  \qquad
  1\le\Keff\le UK\le K .
  \label{eq:keff}
\end{equation}
$\Keff$ is the number of equally used cells that would give the same entropy, also known as the codebook perplexity. For example, a codebook with $K=1024$ and $\Keff=100$ carries as much information per token as a uniformly used vocabulary of 100 words. The bounds follow because entropy is maximal for the uniform distribution on the used cells. $U$ counts every cell that is touched, even rarely, whereas $\Keff$ discounts rarely used cells and can be compared directly across codebook sizes.

%% file: sections/experiments.tex
\section{Experiments}
\label{sec:experiments}

\begin{figure}[t]
  \centering
  \includegraphics[width=\linewidth]{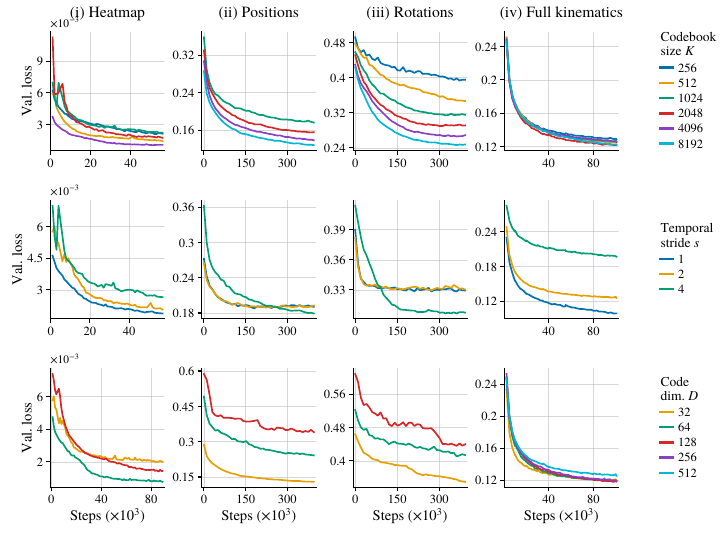}
  \Description{A grid of twelve line plots of validation loss versus training steps. Columns are the four encodings (heatmap, positions, rotations, full kinematics); rows vary codebook size, temporal stride, and code dimension. Larger codebooks lower the loss for positions and rotations; full kinematics curves nearly coincide across codebook sizes and code dimensions; stride 1 is best for heatmaps and full kinematics; code dimension 32 gives the lowest loss for positions and rotations.}
  \caption{Validation loss during training. Columns: encodings (i)--(iv). Rows: codebook size $K$ ($s=2$, $D=32$), temporal stride $s$ ($K=1024$, $D=32$), and code dimension $D$ ($K=1024$, $s=2$). Heatmap losses are scaled by $10^{3}$.}
  \label{fig:val-loss}
\end{figure}

\textbf{Setup.}
For heatmaps, the encoder is a one-layer ResNet followed by a four-layer transformer (8 heads, hidden size 256), and the decoder is a one-layer transformer followed by a linear projection back to the lattice. It is trained with AdamW, a cosine schedule from $10^{-4}$ to $10^{-6}$ with 2000 warm-up steps, 4 epochs, and batch size 32. For (ii)--(iv), the encoder has 16 strided 1D convolutional layers and the decoder 17 layers of nearest-neighbor upsampling and convolution, trained with the same schedule (peak $3\cdot10^{-4}$) for 40 epochs with batch size 64, or 256 for full kinematics. Starting from the default configuration $K=1024$, $s=2$, $D=32$, we vary one factor at a time: $K\in\{256,\dots,8192\}$, $s\in\{1,2,4,8\}$, and $D\in\{32,64,128\}$, with $D\in\{256,512\}$ additionally for full kinematics. \Cref{fig:val-loss} shows validation losses (\cref{eq:vq_loss}), which are not comparable across encodings, and \cref{tab:capacity} the capacity measures of \cref{eq:keff}.

\begin{table}[t]
  \centering
  \caption{Codebook capacity: normalized entropy $\Hn$, utilization $U$ (\%), and effective vocabulary size $\Keff=K^{\Hn}$ (\cref{eq:keff}). Each block varies one factor from the default $K=1024$, $s=2$, $D=32$. Bold: largest $\Keff$ per encoding within a block; --: not run.}
  \label{tab:capacity}
  \scriptsize
  \setlength{\tabcolsep}{2.4pt}
  \begin{tabular}{@{}lr ccr ccr ccr ccr@{}}
    \toprule
    & & \multicolumn{3}{c}{(i) Heatmap} & \multicolumn{3}{c}{(ii) Positions} & \multicolumn{3}{c}{(iii) Rotations} & \multicolumn{3}{c}{(iv) Full kin.} \\
    \cmidrule(lr){3-5}\cmidrule(lr){6-8}\cmidrule(lr){9-11}\cmidrule(l){12-14}
    & & $\Hn$ & $U$ & $\Keff$ & $\Hn$ & $U$ & $\Keff$ & $\Hn$ & $U$ & $\Keff$ & $\Hn$ & $U$ & $\Keff$ \\
    \midrule
    \multirow{6}{*}{$K$}
    & 256  & .48 & 60.16 & 14 & .97 & 92.96 & 217 & .86 & 52.34 & 118 & .95 & 77.78 & 194 \\
    & 512  & .40 & 44.92 & 12 & .95 & 83.78 & 375 & .96 & 87.89 & 399 & .90 & 54.18 & 274 \\
    & 1024 & .30 & 22.75 & 8  & .95 & 85.44 & 724 & .98 & 100.00 & 891 & .84 & 35.14 & 338 \\
    & 2048 & .30 & 14.70 & 10 & .94 & 78.90 & 1296 & .98 & 100.00 & 1758 & .79 & 21.75 & \textbf{413} \\
    & 4096 & .49 & 24.76 & \textbf{59} & .92 & 67.57 & 2106 & .97 & 99.97 & 3191 & .66 & 6.36 & 242 \\
    & 8192 & -- & -- & -- & .89 & 51.85 & \textbf{3040} & .95 & 98.86 & \textbf{5221} & .66 & 4.94 & 383 \\
    \midrule
    \multirow{4}{*}{$s$}
    & 1 & .84 & 35.14 & \textbf{338} & -- & -- & -- & -- & -- & -- & .86 & 97.75 & \textbf{388} \\
    & 2 & .82 & 28.81 & 294 & .98 & 100.00 & \textbf{891} & .98 & 100.00 & \textbf{891} & .30 & 22.75 & 8 \\
    & 4 & .81 & 28.68 & 274 & .95 & 85.44 & 724 & .98 & 100.00 & \textbf{891} & .34 & 15.14 & 11 \\
    & 8 & -- & -- & -- & .75 & 21.97 & 181 & .69 & 13.67 & 119 & -- & -- & -- \\
    \midrule
    \multirow{3}{*}{$D$}
    & 32  & .99 & 97.76 & \textbf{955} & .95 & 85.45 & \textbf{724} & .98 & 100.00 & \textbf{891} & .30 & 22.75 & 8 \\
    & 64  & .99 & 96.05 & \textbf{955} & .73 & 17.29 & 158 & .61 & 8.11 & 69 & .74 & 70.51 & \textbf{169} \\
    & 128 & .93 & 65.69 & 630 & .50 & 3.61 & 32 & .53 & 4.19 & 39 & .44 & 24.22 & 21 \\
    \bottomrule
  \end{tabular}
\end{table}

\begin{figure}[t]
  \centering
  \includegraphics[width=0.72\linewidth]{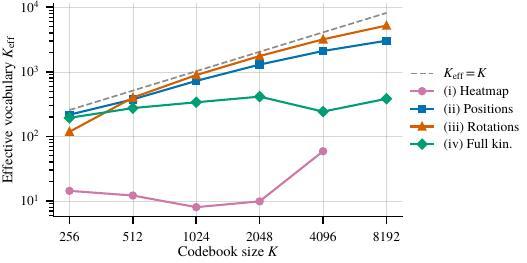}
  \Description{Log-log line plot of effective vocabulary size against codebook size from 256 to 8192, with a dashed diagonal marking full use. Rotations and positions rise close to the diagonal. Full kinematics stays flat between about 200 and 400. Heatmaps stay between about 8 and 59, far below the others.}
  \caption{Effective vocabulary size $\Keff$ against codebook size $K$ (data of \cref{tab:capacity}, $s=2$, $D=32$). The dashed line marks the bound $\Keff=K$, reached when all Voronoi cells are used equally often.}
  \label{fig:keff}
\end{figure}

\textbf{Codebook size.}
Larger codebooks generally lower the validation loss with diminishing returns (\cref{fig:val-loss}, top row); the gain is clearest for rotations and positions and marginal for full kinematics, whose curves nearly coincide. The capacity measures separate the encodings more sharply (\cref{fig:keff}). For positions and rotations, $\Keff$ keeps growing with $K$, by factors of $4.2$ and $5.9$ when $K$ grows from 1024 to 8192, and rotations use virtually every cell ($U\ge98.9\%$) for $K\ge1024$. For full kinematics, $\Keff$ stays between 194 and 413 over a 32-fold range of $K$ while $U$ falls to $4.9\%$: added codewords mostly create Voronoi cells that receive no latents. Since full kinematics has the same per-frame degrees of freedom as rotations (\cref{tab:spaces}), this saturation is likely caused by the redundant parameterization or the auxiliary position loss rather than by the intrinsic dimension of the data. Heatmaps use the fewest cells of all encodings, with $\Keff\le59$ for every $K$.

\textbf{Temporal stride.}
The stride sets how many frames each token summarizes, and hence how long a trajectory segment each Voronoi cell must represent. Its effect depends on the encoding (\cref{fig:val-loss}, middle row). For heatmaps and full kinematics, $s=1$ gives the lowest loss and $s=4$ the highest. Positions converge faster with smaller strides but reach similar final losses, whereas for rotations $s=4$ converges more slowly but ends lowest, so retaining more temporal resolution does not always improve reconstruction. In \cref{tab:capacity}, heatmaps use the most cells at $s=1$, positions and rotations keep $U\ge85\%$ for $s\in\{2,4\}$ but degrade sharply at $s=8$, and full kinematics drops from $\Keff=388$ at $s=1$ to 8 at $s=2$. Since $s=1$ also gives its lowest loss, it is the clear choice for this encoding.

\textbf{Code dimension.}
Lower code dimensions give clearly lower losses for positions and rotations (\cref{fig:val-loss}, bottom row). For full kinematics the final loss is nearly independent of $D$ between 32 and 256, with only $D=512$ noticeably worse, and for heatmaps $D=64$ is best. In terms of capacity, $D=32$ is best or tied for heatmaps, positions, and rotations, while $\Keff$ collapses for positions ($724\to32$) and rotations ($891\to39$) as $D$ grows to 128; full kinematics peaks at $D=64$. A plausible geometric explanation is that the latents concentrate near a low-dimensional subset of $\Sph^{D-1}$, as the low intrinsic dimension of the poses (\cref{tab:spaces}) suggests and as sketched in \cref{fig:schematic}. Extra dimensions then add room for cells that no latent reaches. Testing this hypothesis requires intrinsic-dimension estimates of the latents, which we leave for future work.

\textbf{Practical guidelines.}
Within our setting, the results suggest the following choices for motion tokenizers. For rotation and position encodings, larger codebooks pay off: up to $K=8192$, each doubling of $K$ still adds effective vocabulary, and from $K=1024$ on, rotations keep nearly every cell in use. For the full kinematic encoding, codebooks beyond about 1024 entries mostly add idle codewords, and a temporal stride of $s=1$ is clearly preferable. A code dimension of $D=32$ gives the best utilization for heatmaps, positions, and rotations, and $D=64$ for full kinematics; for positions and rotations, $D=128$ leaves over 95\% of the cells empty. Heatmap tokenizers need particular care: their effective vocabulary never exceeded 59 cells when varying $K$ at $s=2$, so their codebook size should be chosen by $\Keff$ rather than by reconstruction loss alone.

%% file: main.bbl
\begin{thebibliography}{10}
\providecommand{\url}[1]{\texttt{#1}}
\providecommand{\urlprefix}{URL }
\providecommand{\doi}[1]{https://doi.org/#1}

\bibitem{baharani2026feasibility}
Baharani, M., Kimiyaie, A., Alinezhad\_Noghre, G., Danesh\_Pazho, A., Maldonado, G., Tabkhi, H.: A feasibility study on self-supervised llm-inspired training for generalizable human motion understanding. In: 37th British Machine Vision Conference (BMVC) (2026)

\bibitem{vandenboomgaard1994morphological}
van~den Boomgaard, R., Smeulders, A.: The morphological structure of images: The differential equations of morphological scale-space. IEEE Transactions on Pattern Analysis and Machine Intelligence  \textbf{16}(11),  1101--1113 (1994)

\bibitem{10.1007/978-3-030-58571-6_14}
Cai, Y., Huang, L., Wang, Y., Cham, T.J., Cai, J., Yuan, J., Liu, J., Yang, X., Zhu, Y., Shen, X., Liu, D., Liu, J., Thalmann, N.M.: Learning progressive joint propagation for human motion prediction. In: Vedaldi, A., Bischof, H., Brox, T., Frahm, J.M. (eds.) Computer Vision -- ECCV 2020. pp. 226--242. Springer International Publishing, Cham (2020)

\bibitem{dhillon2001concept}
Dhillon, I.S., Modha, D.S.: Concept decompositions for large sparse text data using clustering. Machine Learning  \textbf{42}(1--2),  143--175 (2001)

\bibitem{9879048}
Duan, H., Zhao, Y., Chen, K., Lin, D., Dai, B.: Revisiting skeleton-based action recognition. In: 2022 IEEE/CVF Conference on Computer Vision and Pattern Recognition (CVPR). pp. 2959--2968 (2022). \doi{10.1109/CVPR52688.2022.00298}

\bibitem{tokenhmr}
Dwivedi, S.K., Sun, Y., Patel, P., Feng, Y., Black, M.J.: Tokenhmr: Advancing human mesh recovery with a tokenized pose representation. In: 2024 IEEE/CVF Conference on Computer Vision and Pattern Recognition (CVPR). pp. 1323--1333 (2024). \doi{10.1109/CVPR52733.2024.00132}

\bibitem{alphapose}
Fang, H.S., Li, J., Tang, H., Xu, C., Zhu, H., Xiu, Y., Li, Y.L., Lu, C.: Alphapose: Whole-body regional multi-person pose estimation and tracking in real-time. IEEE Transactions on Pattern Analysis and Machine Intelligence  \textbf{45}(6),  7157--7173 (2023). \doi{10.1109/TPAMI.2022.3222784}

\bibitem{geometric-rigid-nonrigid}
Friji, R., Chaieb, F., Drira, H., Kurtek, S.: Geometric deep neural network using rigid and non-rigid transformations for landmark-based human behavior analysis. IEEE Transactions on Pattern Analysis and Machine Intelligence  \textbf{45}(11),  13314--13327 (2023). \doi{10.1109/TPAMI.2023.3291663}

\bibitem{3692070.3692684}
Geist, A.R., Frey, J., Zhobro, M., Levina, A., Martius, G.: Learning with 3d rotations: a hitchhiker's guide to so(3). In: Proceedings of the 41st International Conference on Machine Learning (2024)

\bibitem{gersho1992vector}
Gersho, A., Gray, R.M.: Vector Quantization and Signal Compression. Kluwer Academic Publishers, Boston (1992)

\bibitem{snapmogen}
guo, c., Hwang, I., Wang, J., Zhou, B.: Snapmogen: Human motion generation from expressive texts. In: Belgrave, D., Zhang, C., Lin, H., Pascanu, R., Koniusz, P., Ghassemi, M., Chen, N. (eds.) Advances in Neural Information Processing Systems. vol. 38, Main Conference, pp. 99939--99955. Curran Associates, Inc. (2025). \doi{10.52202/085713-3341}

\bibitem{momask}
Guo, C., Mu, Y., Javed, M.G., Wang, S., Cheng, L.: Momask: Generative masked modeling of 3d human motions. In: 2024 IEEE/CVF Conference on Computer Vision and Pattern Recognition (CVPR). pp. 1900--1910 (2024). \doi{10.1109/CVPR52733.2024.00186}

\bibitem{humanml3d}
Guo, C., Zou, S., Zuo, X., Wang, S., Ji, W., Li, X., Cheng, L.: Generating diverse and natural 3d human motions from text. In: Proceedings of the IEEE/CVF Conference on Computer Vision and Pattern Recognition (CVPR). pp. 5152--5161 (June 2022)

\bibitem{TM2T}
Guo, C., Zuo, X., Wang, S., Cheng, L.: Tm2t: Stochastic and tokenized modeling for the reciprocal generation of 3d human motions and texts. In: European Conference on Computer Vision (ECCV). pp. 580--597 (2022). \doi{10.1007/978-3-031-19833-5_34}

\bibitem{10377752}
Kong, H., Gong, K., Lian, D., Mi, M.B., Wang, X.: Priority-centric human motion generation in discrete latent space. In: 2023 IEEE/CVF International Conference on Computer Vision (ICCV). pp. 14760--14770 (2023). \doi{10.1109/ICCV51070.2023.01360}

\bibitem{lin2023motionx}
Lin, J., Zeng, A., Lu, S., Cai, Y., Zhang, R., Wang, H., Zhang, L.: Motion-x: A large-scale 3d expressive whole-body human motion dataset. Advances in Neural Information Processing Systems  (2023)

\bibitem{PoseGPT}
Lucas, T., Baradel, F., Weinzaepfel, P., Rogez, G.: Posegpt: Quantization-based 3d human motion generation and forecasting. In: European Conference on Computer Vision (ECCV). pp. 417--435 (2022). \doi{10.1007/978-3-031-20068-7_24}

\bibitem{AMASS:ICCV:2019}
Mahmood, N., Ghorbani, N., Troje, N.F., Pons-Moll, G., Black, M.J.: {AMASS}: Archive of motion capture as surface shapes. In: International Conference on Computer Vision. pp. 5442--5451 (Oct 2019)

\bibitem{7506114}
Mandery, C., Terlemez, {\"O}., Do, M., Vahrenkamp, N., Asfour, T.: Unifying representations and large-scale whole-body motion databases for studying human motion. IEEE Transactions on Robotics  \textbf{32}(4),  796--809 (2016). \doi{10.1109/TRO.2016.2572685}

\bibitem{gsps}
Mao, W., Liu, M., Salzmann, M.: Generating smooth pose sequences for diverse human motion prediction. In: 2021 IEEE/CVF International Conference on Computer Vision (ICCV). pp. 13289--13298 (2021). \doi{10.1109/ICCV48922.2021.01306}

\bibitem{8237550}
Martinez, J., Hossain, R., Romero, J., Little, J.J.: A simple yet effective baseline for 3d human pose estimation. In: 2017 IEEE International Conference on Computer Vision (ICCV). pp. 2659--2668 (2017). \doi{10.1109/ICCV.2017.288}

\bibitem{ICLR2024_e2dd5360}
Mentzer, F., Minnen, D., Agustsson, E., Tschannen, M.: Finite scalar quantization: Vq-vae made simple. In: Kim, B., Yue, Y., Chaudhuri, S., Fragkiadaki, K., Khan, M., Sun, Y. (eds.) International Conference on Learning Representations. vol.~2024, pp. 51772--51783 (2024)

\bibitem{oliveira2026benchmarking}
de~Oliveira, K.C., Brito, I.A., Fernandes, L.A., Rios, W., Filho, A.R.G.: Benchmarking motion tokenizers: A controlled study of human motion quantization. In: ECCV26 Workshop: Human Motion-Informed World Models and Socially Intelligent Action (2026), \url{https://openreview.net/forum?id=J6GFuk0FQR}

\bibitem{vq-vae}
van~den Oord, A., Vinyals, O., Kavukcuoglu, K.: Neural discrete representation learning. In: Advances in Neural Information Processing Systems (NeurIPS). pp. 6309--6318 (2017)

\bibitem{dalle}
Ramesh, A., Pavlov, M., Goh, G., Gray, S., Voss, C., Radford, A., Chen, M., Sutskever, I.: Zero-shot text-to-image generation. In: Meila, M., Zhang, T. (eds.) Proceedings of the 38th International Conference on Machine Learning. Proceedings of Machine Learning Research, vol.~139, pp. 8821--8831. PMLR (18--24 Jul 2021)

\bibitem{serra1982image}
Serra, J.: Image Analysis and Mathematical Morphology. Academic Press, London (1982)

\bibitem{bailando}
Siyao, L., Yu, W., Gu, T., Lin, C., Wang, Q., Qian, C., Loy, C.C., Liu, Z.: { Bailando: 3D Dance Generation by Actor-Critic GPT with Choreographic Memory }. In: 2022 IEEE/CVF Conference on Computer Vision and Pattern Recognition (CVPR). pp. 11040--11049 (Jun 2022). \doi{10.1109/CVPR52688.2022.01077}

\bibitem{9795869}
Sun, Z., Ke, Q., Rahmani, H., Bennamoun, M., Wang, G., Liu, J.: Human action recognition from various data modalities: A review. IEEE Transactions on Pattern Analysis and Machine Intelligence  \textbf{45}(3),  3200--3225 (2023). \doi{10.1109/TPAMI.2022.3183112}

\bibitem{motionbricks}
Wang, T., Dionne, O., De~Ruyter, M., Minor, D., Rempe, D., Zhao, K., Petrovich, M., Yuan, Y., Li, C., Luo, Z., Robison, B., Blackwell, X., Antoniazzi, B., Peng, X.B., Zhu, Y., Yuen, S.: Motionbricks: Scalable real-time motions with modular latent generative model and smart primitives. ACM Trans. Graph.  \textbf{45}(4) (Jul 2026). \doi{10.1145/3811334}

\bibitem{11322811}
Wu, P., Pan, C., Yan, Y., Pang, G., Yan, Q., Wang, P., Zhang, Y.: Deep learning for video anomaly detection: A review. IEEE Transactions on Neural Networks and Learning Systems  \textbf{37}(7),  3010--3030 (2026). \doi{10.1109/TNNLS.2025.3647892}

\bibitem{t2m-gpt}
Zhang, J., Zhang, Y., Cun, X., Zhang, Y., Zhao, H., Lu, H., Shen, X., Shan, Y.: Generating human motion from textual descriptions with discrete representations. In: Proceedings of the IEEE/CVF Conference on Computer Vision and Pattern Recognition (CVPR). pp. 14730--14740 (June 2023)

\bibitem{zhou2019continuity}
Zhou, Y., Barnes, C., Lu, J., Yang, J., Li, H.: On the continuity of rotation representations in neural networks. In: 2019 IEEE/CVF Conference on Computer Vision and Pattern Recognition (CVPR). pp. 5738--5746. IEEE (2019)

\end{thebibliography}
